\pdfoutput=1

\documentclass[11pt]{article}

\usepackage{acl}

\usepackage{times}
\usepackage{latexsym}

\usepackage[T1]{fontenc}

\usepackage[utf8]{inputenc}
\usepackage{xspace}
\newcommand{\method}{{YODA}\xspace}
\newcommand{\teacher}{\textit{teacher}\xspace}
\newcommand{\student}{\textit{student}\xspace}
\newcommand{\basicsimilarharder}{\textit{basic-generalized-harder}\xspace}
\usepackage[linesnumbered,ruled,vlined]{algorithm2e}
\usepackage{amsmath,amsfonts}
\usepackage{microtype}
\usepackage{amsmath} 
\usepackage{inconsolata}
\usepackage{cleveref}
\usepackage{geometry}
\usepackage{multirow}
\usepackage{booktabs}
\usepackage{graphicx, tabularx}
\usepackage{utfsym}
\usepackage[most]{tcolorbox}
\usepackage{pdfpages}
\usepackage{subcaption}
\usepackage{relsize}

\newcommand*\inlinelargeimage[1]{\raisebox{-0.2\baselineskip}{$$\includegraphics[height=1.7\baselineskip]{#1}$$}}

\crefname{section}{§}{§§}
\title{\textsc{YODA}\inlinelargeimage{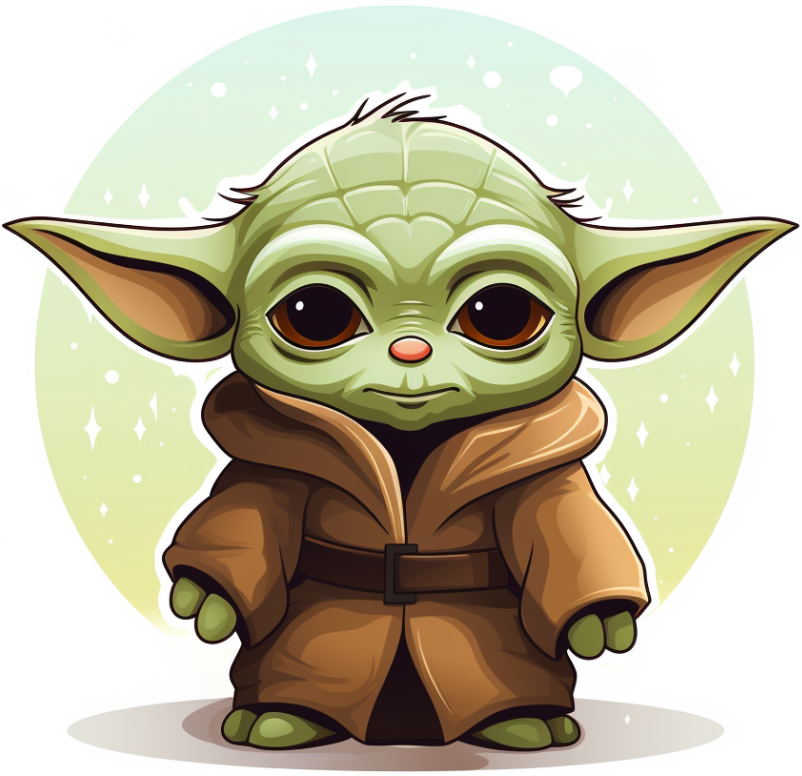}: Teacher-Student Progressive Learning for Language Models}

\author{\textbf{Jianqiao Lu}$^1$\thanks{\hspace{4px}Leading co-authors with equal contribution.}\hspace{4px}\thanks{\hspace{4px}Work done during an internship at Huawei.}~, \textbf{Wanjun Zhong}$^{2*}$\textbf{,}
\textbf{Yufei Wang}$^2$\textbf{,}~\textbf{Zhijiang Guo}$^2$\textbf{,}
\textbf{Qi Zhu}$^{2}$\textbf{,} \textbf{Wenyong Huang}$^{2}$ 
\\
\textbf{Yanlin Wang}$^{3}$\textbf{,} 
\textbf{Fei Mi}$^2$\textbf{,} \textbf{Baojun Wang}$^2$\textbf{,} \textbf{Yasheng Wang}$^2$\textbf{,}  \textbf{Lifeng Shang}$^2$ \textbf{,} \textbf{Xin Jiang}$^2$ \& \textbf{Qun Liu}$^2$
\\
$^1$The University of Hong Kong \ \ \ $^2$Huawei Noah’s Ark Lab \ \ \ $^3$Sun Yat-sen University \\
\texttt{jqlu@cs.hku.hk, zhongwanjun1@huawei.com}\\
}

\begin{document}
\maketitle

\begin{abstract}
\label{sec:abstract}
Although large language models (LLMs) have demonstrated adeptness in a range of tasks, they still lag behind human learning efficiency. 
This disparity is often linked to the inherent human capacity to learn from basic examples, gradually generalize and handle more complex problems, and refine their skills with continuous feedback.
Inspired by this, this paper introduces \method\footnote{Master YODA is a profound educator in the Star Wars series. He is known for his insightful lessons that encourage his students to think critically and look beyond the surface.}, a novel teacher-student progressive learning framework that emulates the teacher-student education process to improve the efficacy of model fine-tuning. 
The framework operates on an interactive \textit{basic-generalized-harder} loop. 
The teacher agent provides tailored feedback on the student's answers, and systematically organizes the education process. 
This process unfolds by teaching the student basic examples, reinforcing understanding through generalized questions, and then enhancing learning by posing questions with progressively enhanced complexity.
With the teacher's guidance, the student learns to iteratively refine its answer with feedback, and forms a robust and comprehensive understanding of the posed questions. 
The systematic procedural data, which reflects the progressive learning process of humans, is then utilized for model training. 
Taking math reasoning as a testbed, experiments show that training LLaMA2 with data from \method improves SFT with significant performance gain (+17.01\% on GSM8K and +9.98\% on MATH). 
In addition, we find that training with curriculum learning further improves learning robustness. 

\end{abstract}
\section{Introduction}
\label{sec:intro}
Large Language Models (LLMs;~\citealt{gpt3.5,instructgpt,chatgpt,gpt4}) have shown remarkable proficiency in a range of tasks. 
Currently, the training of LLMs typically depends on fixed datasets. 
However, these datasets may not be comprehensive enough to cover all aspects necessary for mastering specific skills. 
In contrast, human often acquire skills by effectively extrapolating from limited examples, which offers valuable insights for machine learning.
This human-centric education approach, known as the \basicsimilarharder learning cycle, begins with teaching basic problem-solving skills, then progresses to applying these skills to generalized problems, and gradually advances to more complex problems, which is further strengthened by the ongoing teacher's feedback, allowing students to refine their problem-solving method. 
This indicates a need to evolve the learning strategy for LLMs, pivoting towards an adaptive human-like learning approach that ensures a more systematic and comprehensive exploration and utilization of the available data.

\begin{figure}[t]
    \centering
    \includegraphics[width=0.5\textwidth]{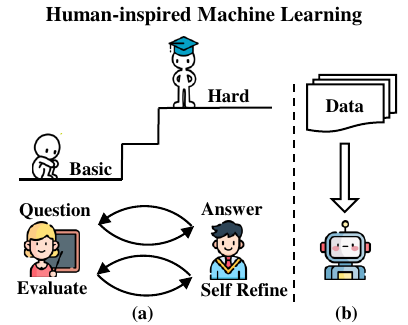}
\caption{Learning paradigms comparison of (a) Human Learning and (b) Machine Learning.}
    \label{fig:motivation}
    \vspace{-0.15in}
\end{figure}

Therefore, we introduce \method, a novel teacher-student progressive learning framework, which imitates the human education process to better explore and extrapolate from limited examples and thereby enhance model learning effectiveness. 
\method operates on a progressive \basicsimilarharder interactive learning loop, which is formed with \student and \teacher agents. 
The \student agent is responsible for iteratively improving its responses based on the \teacher's feedback.
The \teacher agent provides evaluative feedback a   introduces systematically organized questions.
The process is outlined as follows: Initially, the \student begins by tackling a basic problem in the original fixed dataset. Once mastered, the \teacher introduces conceptually similar questions to reinforce the \student's problem-solving skills. 
As proficiency in the current difficulty level is achieved, the \teacher enhances the 
\student's learning by gradually introducing more complex problems, which are further solidified by learning from their generalized variants. 
Throughout this process, the \student iteratively refines their responses to align with the \teacher's evaluation feedback, which enhances learning comprehensiveness. 
Finally, the resulting procedural data from \method provides a systematic and fine-grained reflection of progressive learning and could be utilized for model training. 
This enhances learning effectiveness and mitigates the data scarcity problem by effectively exploring and extrapolating from limited unlabeled problems. 

\begin{figure*}[ht]
    \centering
    \includegraphics[width=\textwidth]{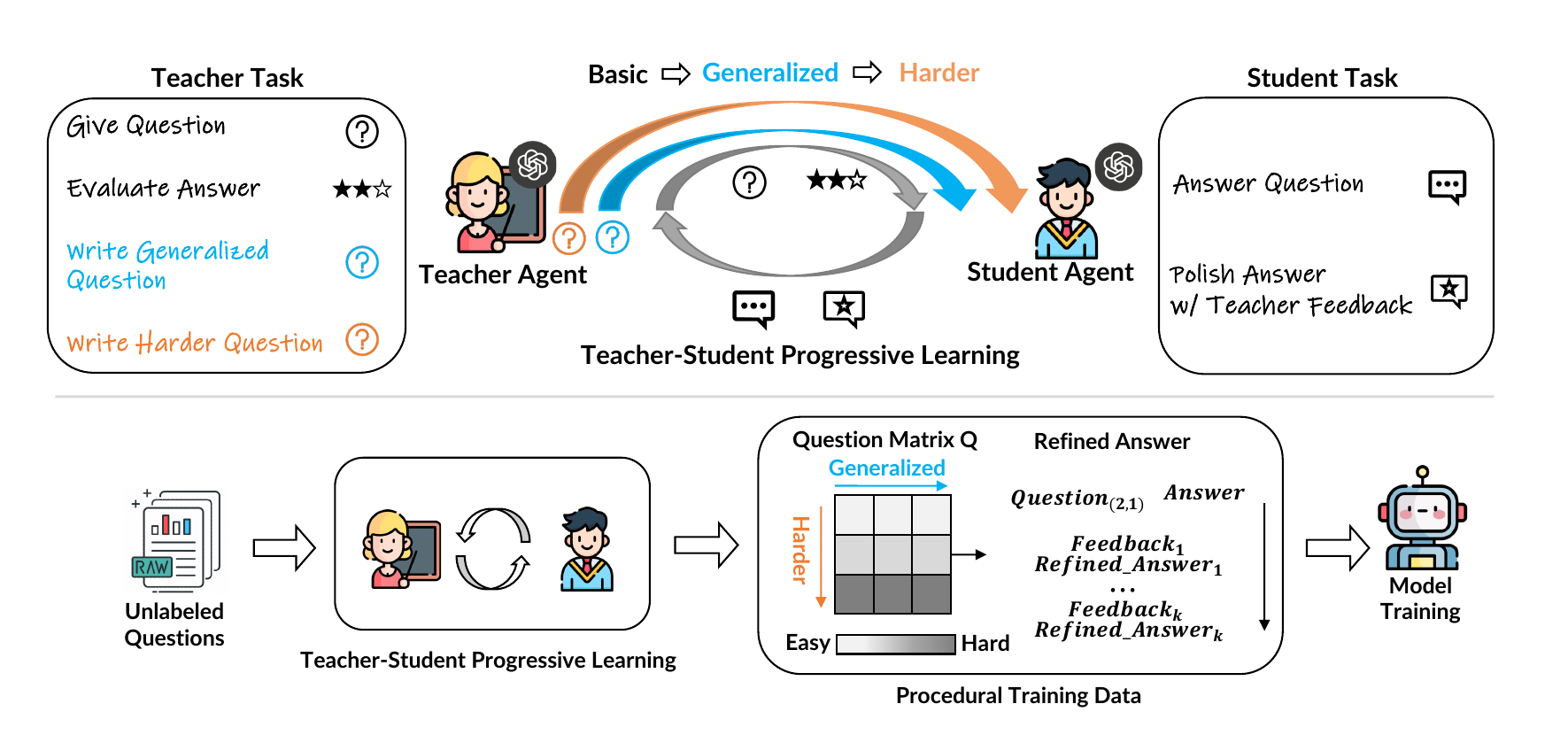}
    \caption{Illustration of the interaction process between the student and the teacher agents. The student agent is tasked with question answering and self-refining its answer with feedback from the teacher agent, who guides the learning by posing generalized/harder questions and evaluating the student's answers.}
    \label{fig:pipeline}
\end{figure*}

To evaluate the efficacy of YODA, we take the math reasoning task as a testbed and train LLaMA2~\citep{llama2} with procedural data from \method.
Experiments show that \method significantly improves standard SFT (+17.01\% on GSM8K and +9.98\% on MATH). 
Further analysis reveals that a training model with curriculum learning that exposes the model to gradually challenging problems also enhances learning robustness.
Moreover, an iterative feedback-refinement cycle is also important in strengthening the learning process. 
These successes highlight that imitating the progressive learning strategy of humans is effective in enhancing model learning effectiveness. 

The key contributions are:
(1) we propose a human-like progressive learning framework \method, enhancing model learning with systematic \basicsimilarharder learning and iterative feedback-refinement.
(2) \method mitigates data scarcity problem with progressively extended data scope, 
(3) Experiments show that \method significantly outperforms baseline models.

\section{Related Works}
\label{sec:related}

\paragraph{Multi-Agent Collaboration}
In the realm of Cooperative Multi-Agent Systems, agents engage in mutual assessment of needs and capabilities, striving for joint actions and knowledge exchange~\citep{Xi2023}. Current LLM-based multi-agent systems largely utilize natural language for communication, which is considered intuitive for human interaction. Pioneering work has explored a dual-agent cooperative system where AI Users issue instructions and AI Assistants provide solutions through interactive multi-turn dialogues, collaborating to execute tasks \citep{LiCamel2023}. Building on this, some research has sought to concentrate this dual-agent cooperation within a singular agent that harnesses both rapid and in-depth thought processes to leverage its strengths~\citep{LinSwiftsage2023}.
\citet{Talebird2023} have proposed a comprehensive LLM-based multi-agent collaboration framework designed to unify individual agents' strengths in a cooperative setting, spawning multiple applications built on this concept~\citep{Liusocial2023,WuAuto2023}.
AgentVerse~\citep{ChenAgentverse2023} advanced this by developing a versatile framework that forms adaptive agent teams responsive to task complexities. To enhance efficiency, some researchers suggest that agents could model successful human cooperation~\citep{ChenSoftware2023}. MetaGPT~\citep{HongMetagpt2023} integrates lessons from the waterfall software development model into agents' input/output standardization, seeking to encode human process management into collaboration prompts. 
These works mainly adopt multi-agent cooperation during inference for complex tasks, while YODA focuses on the training stage and uses multi-agent to stimulate the systematic education process.

\paragraph{Aligning LLMs with Feedback}
Feedback mechanisms are vital for aligning LLMs with human goals, refining their outputs to mirror human values~\citep{Ouyang0JAWMZASR22,Bai2022}. An exemplary method, Reinforcement Learning from Human Feedback (RLHF), tailors LLMs' actions to human preferences without manually defined rewards or direct demonstrations~\citep{Christiano2017DeepRL,Ziegler2019,Bai2022}, making it a popular choice for alignments~\citep{chatgpt,gpt4,llama2}. Despite its effectiveness, RLHF's scalability is hindered by complex setups and resource demands~\citep{ChoshenFAA20,Yuan2023}. New strategies are emerging alongside RLHF. RLAIF replaces human preference labeling with an LLM-based approach~\citep{RLAIF2022}, whereas rejection sampling~\citep{DongRAFT2023} and ReST~\citep{ResT2023} opt for refining responses directly via model-assessed fine-tuning or generative sampling. Methods like HIR~\citep{ZhangLWAG23} and RRHF~\citep{Yuan2023} reframe instruction alignment within goal-oriented reinforcement or response ranking frameworks. Direct Preference Optimization (DPO) and contrastive preference learning offer streamlined or regret-based optimization from preference data~\citep{Rafailov2023,Hejna2023}.  SELF~\citep{lu2023self} combines self-feedback with self-improvement, initiating a meta-skill learning phase for LLMs. Unlike SELF, which uses a single agent for refinement learning, YODA advocates multi-agent collaboration for forming a 
human-like education process to boost learning effectiveness and robustness.

\section{Method}
\label{sec:method}
Our framework draws inspiration from human learning, which progresses from basic examples to increasingly complex problems.
It employs a dual-agent system where a \student agent generates responses and improves upon feedback, while a \teacher agent guides this process by providing new questions that span generalized and more challenging problem scopes and evaluating the student's answers to offer constructive feedback. 

The process is depicted in Fig.~\ref{fig:pipeline} and unfolds in two primary stages: the generation of data through teacher-student progressive learning (\cref{sec:teacher-student-interaction}); and the training of the model using the generated data (\cref{sec:model-tuning}). 
The data generation phase begins with a set of basic problems and evolves through a \basicsimilarharder learning loop~(\cref{sec:progressive-learning}): 
(1) \textbf{Basic}, where the teacher introduces the student to basic problems; (2) \textbf{Generalized}, where the student's understanding is consolidated through questions of comparable difficulty; and (3) \textbf{Harder}, where the teacher challenges the student with increasingly complex problems. 
A key aspect is the \textbf{iterative refinement learning} (\cref{sec:iterative refinement-learning}), where the student answers questions and iteratively refines these answers based on the teacher's feedback. 
This iterative process continues until the teacher deems the answer correct, thereby enhancing learning robustness by teaching the student to self-refine its mistakes.
Finally, the synthesized data collected through the process are utilized for model training, which not only promotes model learning in systematically enhanced question scope, but also solidifies robustness by learning from self-refinement from feedback. 
For a structured overview of this entire workflow, please refer to Algorithm~\ref{alg:nested-learning-refinement}.

\subsection{Teacher-Student Interaction}
\label{sec:teacher-student-interaction}
The teacher-student progressive learning stage forms a systematic learning scope to broaden and deepen the learning scope of the model for reasoning capability acquisition, while the iterative refinement learning process reinforces this knowledge through iterative feedback and self-improvement. 
This combination fosters a systematic and effective learning strategy for the LLMs by making comprehensive utilization and extrapolation from limited basic problems.
All the agent system prompts and task prompts are given in Appendix A.
\subsubsection{Agent Definition}
This section introduces the formal definitions of the \textbf{teacher agent $\mathcal{A}_{teacher}$} and the \textbf{student agent $\mathcal{A}_{student}$}, along with their respective implementation methods.
In our framework, both $\mathcal{A}_{teacher}$ and $\mathcal{A}_{student}$ can be role-played by LLMs with distinct system prompts. 
It's important to note the flexibility in our model selection; $\mathcal{A}_{teacher}$ and $\mathcal{A}_{student}$ can be instantiated using either distinct LLMs or the same one. In our current setup, both roles are acted using the same instance of GPT-4.

\paragraph{Teacher Agent ($\mathcal{A}_{teacher}$)} The primary responsibility of the teacher agent is to test the student agent with a range of questions varying in complexity and generality, and to evaluate the student's responses. The teacher agent's tasks include:

\noindent(1) \textit{Question Generalization} ($QG_{generalized}$): Given a question $q$, the teacher agent generates a set of $m$ new, similarly difficult questions ${q}^i_{generalized}$, applying the prompting function $P_{generalized}$: $\{q^1_{generalized},...,q^m_{generalized}\} = \mathcal{A}_{teacher}(P_{generalized}(q))$.

\noindent(2) \textit{Question Complication} ($QG_{hard}$): For a given question $q^{i-1}_{hard}$, the agent creates a more complex question ${q}^i_{hard}$, increasing the problem's complexity or integrating advanced knowledge, using $P_{hard}$: ${q^{i}}_{hard} = \mathcal{A}_{teacher}(P_{hard}(q^{i-1}_{hard}))$. $i$ indicates the level of difficulty and $q^{1}_{hard}$ is the basic question. 

\noindent(3) \textit{Answer Evaluation} ($Eval$): The teacher agent assesses the correctness of a student's answer $a$ to a question $q$, producing feedback $f$ using $P_{eval}$: ${f} = \mathcal{A}_{teacher}(P_{eval}(q, a))$.

\paragraph{Student Agent ($\mathcal{A}_{student}$)} The student focuses on answering questions and refining its responses based on the teacher's feedback. Its tasks are:

\noindent(1) \textit{Question Answering} ($QA$): The student agent generates an answer $a$ to a given question $q$: $a = \mathcal{A}_{student}(q)$. In cases with a reference example $(\widetilde{q},\widetilde{a})$, the answer is formulated as $a = \mathcal{A}_{student}(q, \widetilde{q}, \widetilde{a})$.

\noindent(2) \textit{Answer Self-Refinement} ($Refine$): Upon receiving the teacher's feedback $f$, the student agent refines its original answer $a$ to question $q$, producing a new answer $r$ using $P_{refine}$: ${r} = \mathcal{A}_{student}(P_{refine}(q, a, f))$.

It's worth discussing that the capabilities of $\mathcal{A}_{teacher}$ need not surpass those of $\mathcal{A}_{student}$. The primary functions of the teacher agent are question recreation and evaluation, and it has been demonstrated that ``evaluation is often easier than task completion'' \cite{chen2023gaining}. Thus, evaluations made by a model of the same caliber can still offer valuable insights for answer refinement \cite{lu2023self, madaan2023selfrefine}.

\subsubsection{Progressive Learning}
\label{sec:progressive-learning}
Given an initial unlabeled basic question corpus  \( U \), the interactive data generation phase in our framework follows a nested \basicsimilarharder loop, structured as follows:

\paragraph{Basic Stage:} Starting with a basic problem \( q \) from the corpus \( U \), $\mathcal{A}_{student}$ generates a response \(a\) using the $QA$ task. Then, $\mathcal{A}_{teacher}$ then assists in mastering basic skills through an iterative evaluation-refinement process, as outlined in \cref{sec:iterative refinement-learning}. This stage lays a robust foundation for further learning.

\paragraph{Generalized Stage:} This stage aims to consolidate the student's reasoning skills. Building upon the basic stage, $\mathcal{A}_{teacher}$ broaden the question scope by presenting $\mathcal{A}_{student}$ with $m$ new questions $ \{q^1_{generalized},...,q^m_{generalized}\} $ that are structurally different but conceptually similar to \( q \).
 Then, $\mathcal{A}_{student}$ learn generalized questions with the same procedure in the basic stage, excluding it can take basic $(q,a)$ as the reference example.

\paragraph{Harder Stage:} 
This phase is designed to enhance the problem-solving skills of $\mathcal{A}_{student}$. The teacher agent challenges the student agent with $n$ increasingly complex questions $\{ {q^1_{hard},...,q^n_{hard}}\} $. Each question $q^i_{hard}$ is an complex version of its predecessor $q^{i-1}_{hard}$. The learning process for these questions mirrors the Generalized Stage, with each $q^i_{hard}$ being further generalized into several generalized questions for solidified learning.

The process creates a detailed question matrix $Q=\{q_{(0,0)},...,q_{(n,m)}\}$, where each question $q_{(complexity, generality)}$ evolves from the original $q_{(0,0)}$ through a series of complication and generalization modifications. In this matrix, $q_{(0,0)}$ is the initial question, $q_{(1,0)}$ is its first complicated version $q^2_{hard}$, and $q_{(1,1)}$ is a generalized variation of $q_{(1,0)}$ with similar complexity. This approach expands the dataset by $(m \times n)$ times, systematically enhancing its complexity and generality, which is crucial for improving the model's problem-solving skills and addressing data limitations. Furthermore, learning of each question in $Q$ undergoes an iterative refinement learning process.

\subsubsection{Iterative Refinement Learning}
\label{sec:iterative refinement-learning}
The iterative refinement learning process enhances the robustness of the learning mechanism by emphasizing the ability to self-refine from mistakes. 

In our model, the student agent ($\mathcal{A}_{student}$) tackles the \(QA\) task by answering each question \(q\) from the question matrix \(Q\). The teacher agent ($\mathcal{A}_{teacher}$) then reviews the response, providing feedback \(f\) through the \(Eval\) task. If the response doesn't meet certain standards, $\mathcal{A}_{student}$ revises the answer (\(r\)) using the feedback, a process known as \(Refine\). This cycle of feedback and refinement continues until the response meets quality criteria or a set number of iterations is reached. This iterative method allows $\mathcal{A}_{student}$ to enhance its understanding and correct errors, thereby increasing its learning effectiveness and adaptability.

\subsection{Model Training}
The teacher-student interaction generates procedural data. 
This section outlines the training strategy involving both crafting training data from the procedural data and training objective, which is crucial for instilling reasoning skills into LLMs. 
\label{sec:model-tuning}

\subsubsection{Procedural Training Data}
\label{sec:multi-process-training-data}
As described in \cref{sec:teacher-student-interaction}, the interactive process generates a diverse set of questions from $Q$, encompassing basic, generalized, and harder problems.
For each $q$, it produces a sequence of intermediate data towards the final revised $r$: $\{q,a,f_0,...,\hat{r_k}\}$, where $k$ denotes the final refinement step.

The total training data $D$ comprises two components: 
(1) \textbf{QA-dataset (\( D_{QA} \))}: Consists of pairs of questions \(q\) and their corresponding final refined responses \(r\), covering the full range from basic, generalized and harder questions.
(2) \textbf{QAFR-dataset (\( D_{QAFR} \))}: Includes detailed tuples of each iteration of self-refinement: (questions \(q\), initial answers \(a_{i-1}\), critiques \(f_{i-1}\), revised responses \(\hat{r_i}\)). This data captures the full iterative refinement process.

This data, rich in diverse question formats and levels of complexity, not only broadens question scope but also reinforces model learning robustness by including the process of mistake correction.

\subsubsection{Training Objective}
The final stage of our framework focuses on leveraging the generated procedural data for training the candidate model $M$. The training objective is structured as follows:
\begin{align}
\label{eq:tuning-process}
\begin{aligned}
   &\mathcal{L}(\phi) = -\left(\mathbb{E}_{(r, q)  \sim D_{QA}} \left[ \log\tau_\phi(r|q) \right] \right. + \\
  & \left. \mathbb{E}_{(r, f, a, q)  \sim D_{QAFR}} \left[  \log\tau_\phi(r|f, a, q) + \log\tau_\phi(f|a, q)  \right] \right)  
\end{aligned}
\end{align}
Here, \(q\) is a question from the matrix \(Q\), \(a\) is the initial response from $\mathcal{A}_{student}$, \(f\) is the feedback provided by $\mathcal{A}_{teacher}$ to \(r\), and \(r\) denotes the refined answer. 
The function \( \tau_\phi(y|x) \) signifies the probability distribution generated by the auto-regressive language model with parameters \( \phi \), predicting the response \( y \) based on the input prompt \( x \).

 This objective is designed to assess the likelihood of $\mathcal{A}_{student}$'s responses being accurate, as well as to evaluate the efficacy of $\mathcal{A}_{teacher}$'s feedback in enhancing $\mathcal{A}_{student}$'s answers. Consequently, the tuning process is oriented towards exposing the model to progressively complex problems, complemented by iterative refinement. 
\begin{algorithm}[h]
\caption{Enhanced Interactive Learning and Refinement Process}
\small
\label{alg:nested-learning-refinement}
  \KwData{Initial question set \(U\)}
  \textbf{Input}: Teacher agent $\mathcal{A}_{teacher}$, Student agent $\mathcal{A}_{student}$, Language Model \(M\)

  \KwResult{Improved Language Model \(M'\)}

  \BlankLine
  \tcp{Define the Generate-Feedback-Refinement Function}
  \SetKwFunction{FIterRefine}{IterRefine}
  \SetKwProg{Fn}{Function}{:}{}
  \Fn{\FIterRefine{\(q,D_{QA}, D_{QAFR}\)}}{
    \tcp{Feedback and Refinement Loop}
    Initialize feedback \(f\) as ``unqualified"\;
    \While{\(f\) is ``unqualified" and within max iterations}{
      Generate response \(a\) for \(q\) using \(\mathcal{A}_{student}\)\;
      Obtain feedback \(f\) from \(\mathcal{A}_{teacher}\)\;
      Refine \(a\) to \(r\) based on \(f\)\;
    }
    Append \((q, r)\) to \(D_{QA}\) and \((q, a, f, r)\) to \(D_{QAFR}\)\;
  }

  \tcp{Iterative Learning Process}
  Initialize \(D_{QA}\), \(D_{QAFR}\), \(Q_{1} = U\)\;
  \For{\(i = 1\) to \(n\)}{
    \tcp{Progressive Question Generation}
    Initialize \(Q_{i+1}\) as empty\;
    \ForEach{\(q\) in \(Q_{i}\)}{
      \FIterRefine{\(q, D_{QA}, D_{QAFR}\)}\;
      \tcp{Generate and Process Generalized Questions}
      \For{\(j = 1\) to \(m\)}{
        Generate \(q_{generalized}\) from \(q\)\;
        \FIterRefine{\(q_{generalized}, D_{QA}, D_{QAFR}\)}\;
      }
      \tcp{Generate Harder Question}
      Generate \(q_{harder}\) from \(q\)\;
      Add \(q_{harder}\) to \(Q_{i+1}\)\;
    }
  }

  \tcp{Model Training with Refined Data}
  Define objective as in~\cref{eq:tuning-process}\;
  \(M'\) = Supervised Fine-Tuning of \(M\) using \(D_{QA} \cup D_{QAFR}\)\;

  \tcp{Complete Training}
  \Return Enhanced Language Model \(M'\)\;
\end{algorithm}

\section{Experiments}
\label{sec:experiment}
Our experiments aim to answer the following questions:

\noindent 1. How does the performance of \method compare to other baselines?  (\cref{sec:main-result})  
     
\noindent 2. How do the features of iterative refinement, generalized stage, and harder stage contribute to the effectiveness of 
 \method? (\cref{sec:ablation}) 
 
\noindent 3. Does curriculum learning, by following an easy-to-hard progression, lead to more effective use of training data?  (\cref{sec:curriculum-learning})
 
\noindent  4. How does the amount of training data used impact the performance of \method (\cref{sec:data-size-effect})







\subsection{Experimental Setup}
\label{sec:main-experiment}
\paragraph{Datasets.} 
We select mathematics as our testbed since it closely mirrors the way humans think and solve problems. 
This allows us to effectively evaluate how well LLMs can reason and tackle complex problems.
Our experiment utilizes the following benchmarks \textbf{GSM8K}~\citep{Cobbe2021} and \textbf{MATH}~\citep{HendrycksBKABTS21}.
The \textbf{GSM8K} dataset comprises a diverse collection of high-quality grade school math problems, totaling approximately 7.5K for training and 1K for testing. These problems typically require 2 to 8 reasoning steps and encompass a wide range of complexity levels, ensuring a balanced evaluation. The \textbf{MATH dataset} is composed of high school-level competition problems in areas such as Algebra, Geometry, and Precalculus. It presents a greater challenge than GSM8K, comprising 7.5K training problems and 5K for testing, and demands more advanced mathematical reasoning skills.
\paragraph{Implementation and Evaluation}
In our implementation, GPT-4 is utilized as both the student agent (\(\mathcal{A}_{student}\)) and the teacher agent (\(\mathcal{A}_{teacher}\)). This approach significantly enhances the quality and efficiency of procedural data generation. The high-quality data generated is then used to fine-tune a smaller foundational model, specifically LLaMA2-7B, ensuring a more effective and cost-efficient training process.
For additional training details and used prompts, please refer to Appendix~\cref{app:training-details} and~\cref{app:interplay-of-roles}.
During inference, our objective is to directly assess the model's capabilities. To achieve this, we employ the straightforward Question-Answering ($QA$) task (\cref{sec:teacher-student-interaction}) to extract answers from the model using a Zero-shot Chain of Thought (CoT) methodology~\citep{KojimaGRMI22},
The efficacy of our model is evaluated based on the accuracy of the final computational results.

\paragraph{Baselines.}
As our objective is to improve the learning strategy of LLMs, we compare \method with the following baselines. More implementation details can be found at \cref{app:data-information-training-methods}.

\noindent(1) \textbf{Supervised Fine-Tuning (SFT)} fine-tunes models using human-labeled, ground truth data.

\noindent(2) \textbf{AI-Supervised Fine-Tuning (AI-SFT)} involves training the model with data instances generated by GPT-4. Each instance comprises an unlabeled prompt sourced from ground truth and the corresponding direct response produced by GPT-4.

\noindent(3) \textbf{Reinforcement Learning from Human Feedback (RLHF)}~\cite{ChristianoLBMLA17} uses reinforcement learning with a reward model trained on human preference data to align LLMs.
RLHF applys the SFT model from (1).
The reward model is trained by ranked responses from \(D_{QAFR}\), assuming the refined answer \(r\) is better than \(a\).

\noindent(4) \textbf{Chain of Hindsight (CoH)}~\cite{HaoLiuCOH2023}: 
To help the model in learning preference via SFT, CoH separates the good and bad examples with indicators in SFT data. 
We take refined answer~\(r\) from \(D_{QAFR}\) as a positive example while treating  \(a\) as the negative one. The model is then inferred with the ``positive'' indicator. 


\noindent(5) \textbf{Self-Evolution with Language Feedback (SELF)}~\cite{lu2023self} adopts self-evolution training to enhance LLM learning. 
The model training initially follows the approach outlined in Eq.~\ref{eq:tuning-process}. The initial response \(a\) is produced by LLaMA2, and both the feedback \(f\) and the refined response \(r\) are supplied by GPT-4. Subsequently, the model undergoes single round of self-evolution.

\noindent(5) \textbf{WizardMath}~\cite{WizardMath} augments math prompts in two ways: it either simplifies questions or creates more challenging ones. The results presented are directly taken from the original paper.

\subsection{Main Result}
\label{sec:main-result}
\begin{table}[ht]\scriptsize
    \centering

     \scalebox{1.25}{
    \begin{tabular}{l|cc}
    \toprule
    Models&\textbf{GSM8K}&\textbf{MATH}\\
    \midrule
    \midrule
    \multicolumn{3}{c}{\textbf{Other Methods}}\\
    \midrule
      LLaMA2 + RLHF& 40.12 & 7.28  \\
      LLaMA2 + COH& 43.41 & 8.14 \\
      LLaMA2 + SELF& 44.31  & 8.32 \\
      LLaMA2 + WizardMath & 54.90 & 10.70  \\
    
     \midrule 
    \multicolumn{3}{c}{\textbf{Direct Baselines}}\\
     \midrule
     LLaMA2 & 11.67 & 1.80 \\
     LLaMA2 + SFT& 39.27 & 6.22   \\
     LLaMA2 + AI-SFT  &  43.06 &  7.12   \\

     \midrule
     \multicolumn{3}{c}{\textbf{Our Methods}}\\
     \midrule
     \method & 60.07 & 17.10  \\
     \method + GT-filter & 61.13 & 17.60 \\
    
     \bottomrule
    \end{tabular}
    }\caption{\label{tab:main-result}Comparison of End-to-End Results: Our approach and baseline methods were compared on the GSM8K and Math datasets. 
    The results presented are based on the zero-shot cot for all methods evaluated.}
\end{table}
Table \ref{tab:main-result} presents a comparison between the performance of YODA and other baseline models, revealing several key findings.

\paragraph{YODA significantly enhances SFT.} 
The primary objective of YODA is to improve the learning strategy beyond typical SFT with fixed data. 
As shown in Table \ref{tab:main-result}, using the same set of seed prompts, the \method framework achieves a marked improvement over its direct baseline model LLaMA2 + AI-SFT ($43.06\%\xrightarrow{+17.01\%}60.07\%$ on GSM8K and $7.12\%\xrightarrow{+9.98\%}17.10\%$ on MATH). Notably, \method also significantly outperforms its counterpart (SFT) trained with human-labeled ground-truth data. 
These findings highlight the efficiency of \method in optimizing the use of existing data via a progressive learning approach, significantly boosting the learning performance of LLMs.

\paragraph{YODA as an effective learning strategy.} 
To further evaluate YODA's efficacy, we compare it against RLHF and CoH, which are established methods for improving LLMs. 
Although Chain of Hindsight (CoH) shows slight gains over its baseline AI-SFT, and RLHF fails to exhibit notable progress, YODA consistently surpasses both methods in performance. 
The restrained success of RLHF can be largely attributed to the difficulty in accurately assessing the quality of correct reasoning chains in math problems using a single scalar reward.
\method proves to be a more effective learning approach by mimicking human-like learning processes and employing detailed, informative language feedback to guide the iterative refinement process.

\paragraph{Advantage of teacher-student learning over single-agent learning.}
While SELF, which employs training based on refined self-generated responses, shows improvement over the baseline LLaMA2 model, it falls short of the performance achieved by the YODA framework. This distinction underscores the advantages of a teacher-student learning approach as compared to the self-evolutionary process of a single agent, like in SELF. The teacher-student interaction facilitates a broader exploration of data and fosters a more systematic and structured learning process. 
This, in turn, significantly enhances the model's ability to grasp and apply complex concepts.

\paragraph{The Advantage of Progressive and Refinement Learning.} The \method approach, with its performance of 60.07\% on the GSM8K dataset and 17.10\% on the MATH dataset, differs significantly from WizardMath, which achieved 54.90\% and 10.70\% respectively. 
This difference in results stems from distinct methodologies. \method  facilitates the transition from simpler to more complex problems and aids in generating generalized questions for each problems. 
In contrast, WizardMath focuses on prompt complication to expand the question exploration space, and it does not incorporate the feedback-refinement training data that \method uses.

\paragraph{Enhanced performance with high-quality data.} 
Incorporating a ground-truth label filtering process in the basic stage of training (Appendix \ref{app:filtering}), YODA + GT-filter enhances the quality and reliability of the training data. This approach yields a slight performance improvement, with scores reaching 61.13\% on GSM8K and 17.60\% on MATH. These results demonstrate that the quality of training data is a crucial factor in model training. Furthermore, they suggest that the data from \method is inherently of high quality, as evidenced by the relatively modest gains.




\subsection{Ablation Study}
\label{sec:ablation}
\paragraph{Settings.}
To bring in-depth analysis about the functionality of main components in \method (i.e., Iterative refinement, Generalized Stage, Harder Stage), we conduct ablation studies. The settings are introduced below. 

\noindent (1) \textbf{YODA w/o Refinement}: For each question in \(Q\), the training data \(D_{QA}\) is formed by using only the \(QA\) task to generate the initial answer pair \((q,a)\), without including any refinement data from \(D_{QAFR}\).

\noindent (2) \textbf{YODA w/o Generalized Stage}: 
The model's exposure is limited to a subset of questions with increasing complexity.
We only use the questions from the first column ($\{q_{(0,0)},...,q_{(n,0)}\}$) in $Q$ with increased complexity, and exclude generalized questions. 

\noindent (3) \textbf{YODA w/o Harder Stage}: 
We only use the data from the first row ($\{q_{(0,0)},...,q_{(0,m)}\}$) in the question matrix $Q$, which encompasses only those questions that are generalized to the basic questions. 

\begin{table}[ht]
\centering
   
 \scalebox{0.85}{ 
\begin{tabular}{ccc|cc}
\toprule
 \textit{Refinement} & \textit{Generalized} & \textit{Harder} &  \textbf{GSM8K} & \textbf{MATH} \\ 
\midrule
\usym{1F5F8}  & \usym{1F5F8} & \usym{1F5F8}   &  \textbf{60.07} &   \textbf{17.10}  \\ 
\usym{1F5F8}& \usym{1F5F8}& \usym{1F5F4}&56.51&14.86 \\
\usym{1F5F8}& \usym{1F5F4}& \usym{1F5F8}&55.45&14.40\\
\usym{1F5F4}& \usym{1F5F8}& \usym{1F5F8}&57.57&15.16\\
\usym{1F5F4}& \usym{1F5F4}& \usym{1F5F4}&43.06&7.12\\
\bottomrule
\end{tabular}
}\caption{\label{tab:ablation-result-similar-evolve-qa-self}Performance metrics for different training components in GSM8K and MATH  benchmarks.}
\end{table}
\paragraph{Analysis.}
From this experiment, we highlight the following findings:
(1) Eliminating refinement leads to -3.50\% and -1.94\% absolute performance drop on GSM8K and MATH respectively, showing that learning self-refinement is critical in enhancing the learning effectiveness and robustness, which is consistent with SELF~\cite{lu2023self}.
(2) Removing generalized stage drops the performance by -4.62\% on GSM8K and -2.70\% on MATH, demonstrating that learning with generalized questions is critical in consolidating the problem-solving capability of LLMs.
(3) Ablation of harder stage brings observed performance drop (-3.56\% on GSM8K and -2.24\% on MATH). This illustrates the critical role of progressively increasing problem complexity in systematically advancing the capabilities of LLMs.
(4) 
The comprehensive ablation of all components results in a marked performance decline (-17.01\% on GSM8K and -10.72\% on MATH), strongly evidencing the effectiveness of the progressive learning process in enhancing the learning strategy of LLMs.




\subsection{Curriculum Learning Analysis}
\label{sec:curriculum-learning}
Curriculum learning \citep{soviany2022curriculum} is a training methodology that trains models from simpler tasks to increasingly complex ones. 
To implement curriculum learning, we arrange the training data following the structure in Alg.~\ref{alg:nested-learning-refinement}, where $i$ denotes the complexity level.
We divide the training into three phases: \textbf{basic}, \textbf{intermediate}, and \textbf{advanced}, each corresponding to a specific complexity level \(i\). 
In each stage, the training data is organized using a weighted sampling method, consisting of 80\% data from the current complexity level and 20\% from the other levels.
Subsequently, data subsets of increasing complexity are progressively integrated into the model's training.
We compare curriculum learning with vanilla learning, where all data is randomly shuffled. 

\begin{figure}[ht]
    \centering
\includegraphics[width=0.5\textwidth]{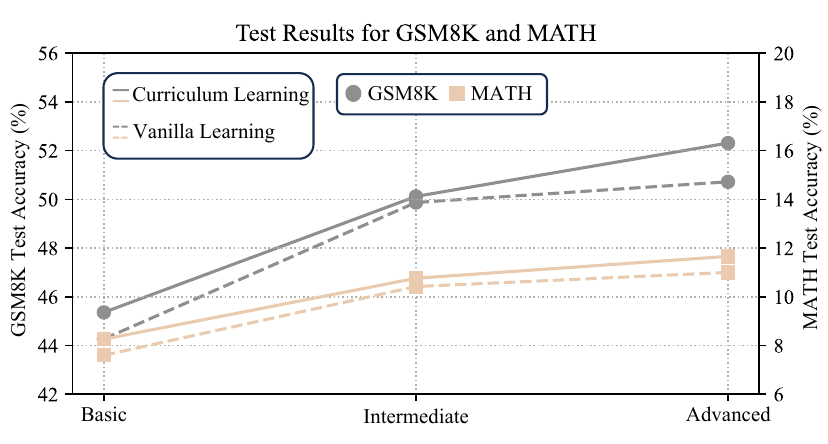}
     \caption{Comparative test results of curriculum learning on GSM8K and MATH test sets.}
     \label{fig:curriculum-learning-test-results}
\end{figure}

As the experiment results\footnote{The best result in Figure~\ref{fig:curriculum-learning-test-results} is lower than the one in Table~\ref{tab:main-result} because the model is only trained for one epoch to eliminate the influence on data repetition, while the full version uses 3 epochs. } shown in Figure~\ref{fig:curriculum-learning-test-results}, curriculum learning derives robust learning curves that continually improve performance and yield better final performances compared with vanilla learning on both datasets. 
This systematic progression from basic to more challenging training data mirrors the human progressive learning process. It underscores the effectiveness of a structured approach in enhancing the model's capacity to build upon foundational knowledge and master more advanced skills with greater efficacy.


\subsection{Data Size Effect}
\label{sec:data-size-effect}
\begin{figure}[ht]
    \centering
    \includegraphics[width=0.5\textwidth]{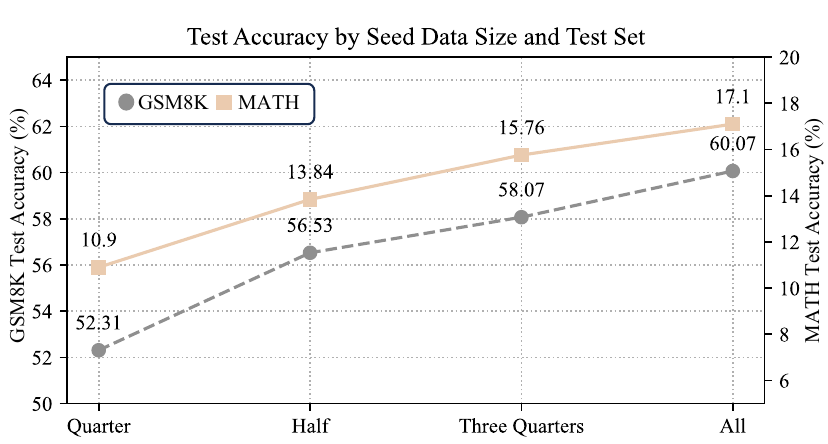}
    \caption{
    Performance of YODA on GSM8K and MATH testsets with different sizes (quarter, half, three-quarters, and full data) of seed basic problems. }
    \label{fig:data_size_effect}
\end{figure}

Given the pivotal role of data scale in LLM training, we analyze \method's performance using varying amounts of seed basic problems. As illustrated in Figure \ref{fig:data_size_effect}, our findings consistently indicate that a larger scale of seed problems leads to enhanced model performance across both datasets. Notably, when the training data scale constitutes less than half of the total, we observe a substantial performance boost with increased seed problems. Subsequently, performance exhibits a gradual ascent at a lower speed as data volume continues to grow. 
Our primary findings can be summarized as follows: 1. The impact of data quantity on performance is distinctly noticeable. 2. The trend observed in the curve suggests that further increasing the volume can enhance performance. 


\section{Conclusion}

This paper introduces \method, a teacher-student progressive learning framework that emulates the interactive education process inspired by interactive human education processes, aimed at boosting the efficiency of model learning. 
\method implements a \basicsimilarharder beginning with basic problem-solving and gradually addressing increasingly complex challenges. 
Through an iterative process guided by a \teacher agent, the \student agent refines its responses based on feedback, fostering a deep and thorough understanding of the questions.
The framework generates procedural data that is then employed in model training. This approach effectively overcomes data limitations by more effectively utilizing available data. It leads to a notable improvement in the effectiveness of LLMs, creating a structured learning journey similar to human education systems.
Our experiments on mathematical benchmarks reveal that \method markedly enhances the baseline performance, achieving a 17.01\% absolute improvement on GSM8K and a 9.98\% increase on MATH. 
Additional analysis suggests that the integration of curriculum learning into \method further strengthens the model's learning capabilities.
\label{sec:conclusion}
\section{Limitation}
\label{sec:limitation}

\bibliography{custom}

\appendix

\section{Appendix}
\label{sec:appendix}

\subsection{Training Details }
\label{app:training-details}
\paragraph{Training Hyperparameters}

Our experiments were carried out using 8 V100 GPUs, each with 32GB of memory. 
The training hyperparameters employed are detailed in Table~\ref{tab:training-parameters} . 
These settings remained the same for all experiments, except for those discussed in~\cref{sec:curriculum-learning}.
\begin{table}[h]
\caption{Training Hyperparameters}
\centering
\begin{tabular}{cc}
\toprule
\textbf{Hyperparameter} & \textbf{Value} \\
\midrule
Global Batch Size & 128 \\
Learning Rate & \(2 \times 10^{-5}\) \\
Epochs & 3 \\
Max Length & 2048 \\
Weight Decay & 0 \\
\bottomrule
\end{tabular}
\label{tab:training-parameters}
\end{table}

\paragraph{Training data}
In \cref{alg:nested-learning-refinement}, the initial question set \(U\) comprises unlabeled prompts from GSM8K and MATH training datasets. The \textit{IterRefine} function operates with a maximum of 3 iterations. The algorithm advances through 3 levels of difficulty (\(n = 3\)), with each level \(Q_{i}\) involving the generation of 2 additional generalized questions (\(m = 2\)) for every original question. This results in 3 different difficulty levels, with 3 generalized questions generated at each level. Each question is subjected to the IterRefine process up to 3 times.
The collected data, comprising a total of 170K instances that include both question answers and the refinement process, is utilized for fine-tuning LLMs.

\subsection{Related Works}
\paragraph{Large Language Model for Math}
Early works usually attempted to pre-train specific models for solving mathematical problems~\citep{Taylor2022, LewkowyczADDMRS22}, based on the carefully selected mathematical
corpus. Since the reasoning capability of the language model can significantly benefit from the general corpus like text and code, some works also choose to fine-tune the general language model to elicit mathematical abilities~\citep{LuoWizard2023,YueMammoth2023,YuMeta2023,Azerbayev2023}.
Recently, LLMs have shown great success in mathematical reasoning, with comparable accuracy compared to fine-tuned baselines, simply by providing a few examples of problem-solving processes.  Early success includes reasoning with a step-by-step chain of thought (CoT;~\citealt{Wei0SBIXCLZ22}), decomposing questions into sub-questions in a least-to-most fashion~\citep{ZhouSHWS0SCBLC23}, zero-shot prompting LLMs with simply one sentence~\citep{KojimaGRMI22}, writing programs to solve procedural tasks~\citep{GaoMZ00YCN23,ChenPoT2022}.
Despite generating solutions in a single forward pass, one line of work employs multiple reasoning results and ensembles them by majority vote~\citep{WangWSLCNCZ23}, retrieves exemplars based on CoT dual queries~\citep{Xiong2023} and stepwise verifier~\citep{LiStep2022}.

\subsection{Filtering Details}
\label{app:filtering}

To enhance the quality of our training dataset, we implemented a GT-filter post-processing step. 
The GT-filter process involved reviewing the output of GPT-4 and selectively retaining only those data points in the basic stage that matched the ground truth. 
This filter refined the dataset by leveraging ground truth to ensure maximum accuracy. Prior to applying the GT-filter, the accuracy of the training data generated by GPT-4 was 84.34\% for GSM8K and 35.4\% for the MATH dataset.

The GT-filter greatly enhanced the training data's reliability, contributing to the performance gains shown in~\cref{tab:main-result}.

\subsection{Interactive Teacher-Student Evaluation Refinement}
\label{app:interactive-refinement}

Our model introduces a response refinement process that significantly enhances data accuracy. During the basic stage, solely relying on GPT-4 generated responses resulted in an initial accuracy of 81.82\% for GSM8K and 31.2\% for the Math dataset. However, by employing an interactive teacher-student evaluation refinement, the accuracy improved to 84.34\% for GSM8K and 35.4\% for Math, demonstrating a noticeable enhancement.

This refinement process entails a teacher-student system collaboratively evaluating and refining the responses from GPT-4. Through this dynamic interaction, the system identifies and corrects inaccuracies, thereby optimizing the quality of the dataset used for training our model.
This underscores the efficacy of the interactive teacher-student evaluation refinement in our model's training regime, bolstering the dataset's accuracy and the model's performance.

\subsection{Interplay of Roles}
\label{app:interplay-of-roles}
Within the framework of \method, we have established distinct roles for agents to accurately simulate the interactive dynamics of human learning. These roles are clearly delineated through a series of prompts, which guide the actions and responses of each agent. Presented below are two separate sets of prompts, specifically designed for the \student and \teacher agents.

\subsubsection{Teacher Prompts}
\textbf{Teacher System Prompt} outlines the guidelines for evaluating \student answers in the subject of mathematics.
\begin{tcolorbox}[title=Teacher System Prompt]
You are a professional teacher teaching math subject. 

Given a math problem, your task is to evaluate the answer from the student. 

If ask, you can test the student with newly generalized question or newly generated harder question.
\end{tcolorbox}
\noindent\textbf{Teacher Evaluation Prompt}~focuses on providing critiques and scores for \student's responses to math questions.
\textit{``For the question: \{question\},
here is a proposed answer:
\{answer\}
Provide feedback or critique for the previous response. 
Please keep it under 100 words. 
Rate the quality of the answer on a scale from 1 (being poor or nonsensical) to 10 (perfect). 
Also indicate whether a revision is necessary by stating either ``Revision is needed'' or ``Revision is not needed''.
Typically, a score below 9 suggests that a revision should be considered.
Feedback:''}

\noindent\textbf{Teacher Questioning Prompt}~includes two specific tasks for the teacher:

\noindent(1)The ``Generalized Question'' task involves creating a new question inspired by a given one.
\textit{``Your task is to use the provided question as a basis for crafting a new, but generalized question. This newly formulated question should require a generalized approach to find the solution but should be phrased differently. Ensure that the new question is of comparable length and complexity to the original. It should be clear, concise, and answerable by people.
Given Question: 
\{question\}
Created Question: ''}

\noindent(2)The ``Harder Question'' task entails rewriting a given question to increase its complexity.
\textit{``Your role is to modify an existing math question, elevating its complexity to challenge advanced AI systems. 
To achieve this, incorporate an additional constraint or requirement into the original question. 
The modified question should be more challenging, requiring more intricate solutions or additional mathematical steps. 
Strive to keep the question concise and understandable to ensure it remains solvable by humans.
Given Question: 
\{question\}
Created Question: ''}

\subsubsection{Student Prompts}
\textbf{Student System Prompt} provides instructions for \student's on how to respond to math questions.
\begin{tcolorbox}[title=Student System Prompt]
As a student studying mathematics, your responsibility is to thoughtfully respond to math questions. If you receive feedback from your teacher, use it to revise and improve your answers. This approach helps you learn and understand mathematical concepts more deeply.
\end{tcolorbox}
\textbf{Student Response Prompt} offers a structured approach for \student's to answer math questions step by step.
\textit{``Answer the question: \{question\}. 
Let's think step by step.''}
\textbf{Student Refinement Prompt} is designed for \student's to refine their answers based on \teacher feedback, focusing on clarity and simplicity.\textit{``When provided with a question, your initial answer, and feedback from a teacher, your task is to revise the answer accordingly. Focus on incorporating the teacher's feedback to improve your response. Present the revised answer clearly and concisely, as if it were your first attempt at answering the question.
The input question is \{question\}
The original answer is \{answer\}
The feedback to the answer is \{feedback\}
Output revised answer:''}

\subsection{Detailed Data Information for Training Methods}
\label{app:data-information-training-methods}
\noindent(1)~\textbf{SFT}: This method fine-tunes models using 15K human-labeled, ground truth data instances of GSM8K and MATH.

\noindent(2)~\textbf{AI-SFT}: Involves training the model with 15K data instances generated by GPT-4. Each instance comprises an unlabeled prompt sourced from ground truth and the corresponding direct response produced by GPT-4.

\noindent(3)~\textbf{RLHF}: RLHF uses only the basic stage data from \method,  
We implement this using the trlx framework\footnote{https://github.com/CarperAI/trlx}. The reward model is trained with 6K pairs of ranked responses from  \(D_{QAFR}\).

\noindent(4)~\textbf{COH}: We have implemented the COH method according to its official guidelines\footnote{https://github.com/lhao499/chain-of-hindsight}.
In our approach, the training dataset for CoH consists of 12K instances sourced from \(D_{QAFR}\) generated during the basic stage of \method.
These instances are categorized into two categories: ``positive only'' (p) and ``negative-positive'' (np), following the official guidelines.
The refined answers \(r\)  are treated as positive examples, while the initial responses \(a\) are considered negative examples. 

\noindent(5)~\textbf{SELF}: The training dataset consists of 21K instances, including 15K QA data entries and 6K QAFR data entries. The initial responses are produced by LLaMA2, while the feedback and refined responses are provided by GPT-4.


\end{document}